Yan Zhang [ID], Xiao He [ID], Shuanhu Gao [ID], and Aimin Zhou [ID]
*East China Normal University, CHINA*

Hao Hao [ID]
*Shanghai Jiao Tong University, CHINA*

# Evolutionary Retrosynthetic Route Planning

## Abstract

Molecular retrosynthesis is a significant and complex problem in the field of chemistry, however, traditional manual synthesis methods not only need well-trained experts but also are time-consuming. With the development of Big Data and machine learning, artificial intelligence (AI) based retrosynthesis is attracting more attention and has become a valuable tool for molecular retrosynthesis. At present, Monte Carlo tree search is a mainstream search framework employed to address this problem. Nevertheless, its search efficiency is compromised by its large search space. Therefore, this paper proposes a novel approach for retrosynthetic route planning based on evolutionary optimization, marking the first use of Evolutionary Algorithm (EA) in the field of multi-step retrosynthesis. The proposed method involves modeling the retrosynthetic problem into an optimization problem, defining the search space and operators. Additionally, to improve the search efficiency, a parallel strategy is implemented. The new approach is applied to four case products and compared with Monte Carlo tree search. The experimental results show that, in comparison to the Monte Carlo tree search algorithm, EA significantly reduces the number of calling single-step model by an average of 53.9%. The

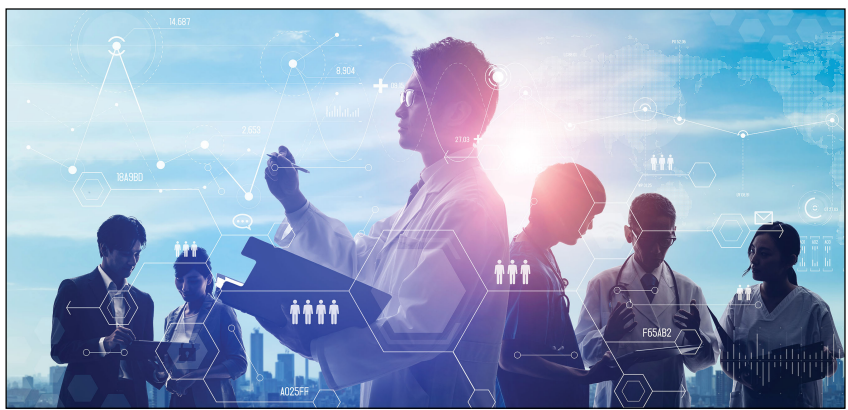
©SHUTTERSTOCK/METAMORWORKS

time required to search three solutions decreases by an average of 83.9%, and the number of feasible search routes increases by 1.38 times. The source code is available at https://github.com/ilog-ecnu/EvoRRP.

## I. Introduction

Molecular retrosynthesis is of great importance in various fields [1], [2], including drug synthesis and catalyst design. It enables rapid determination of potential pathways and starting materials for the synthesis of target compounds, effectively providing essential guidance for synthesis routes. Retrosynthesis contains two types of process: single-step retrosynthesis and multi-step search process. The single-step retrosynthesis process involves breaking down an organic molecule into its original reactants. As illustrated in Figure 1, the retrosynthetic route for target product is generated using this single-step method. Moreover, the intermediate molecules, $R^1$ and $W^1$, are obtained through the single-step model with the input target product $P$. $W^1$ belongs to the set $\Psi$, which represents the building block dataset containing commercially available molecules, serving as the terminal reactant database. The process continues by selecting $R^1$ as input for the single-step model, and this cycle repeats until all products belong to $\Psi$, forming a multi-step retrosynthetic reactions.

In recent years, single-step model, particularly used in architectures like transformers, has played a crucial role in advancing this field. A transformer architecture [3] has become a common choice and practical solution for single-step retrosynthetic tasks. It has achieved significant advancements and greatly improved search effectiveness, leading to remarkable progress in the field of retrosynthesis.

This article was recommended for publication by Associate Editor Rong Qu. Corresponding author: Aimin Zhou (e-mail: amzhou@cs.ecnu.edu.cn).

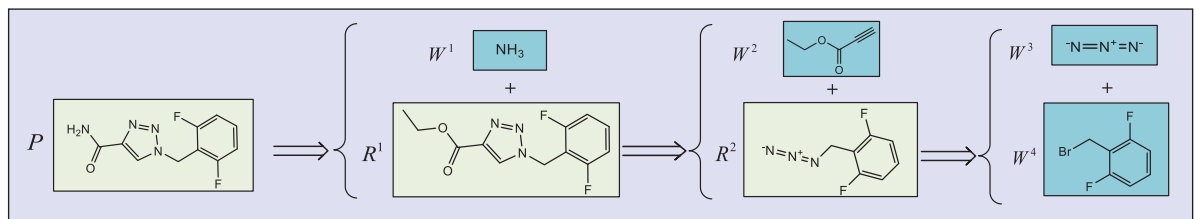

**FIGURE 1** Multi-step retrosynthetic route from target product $P$ to ingredients $W$. The single-step retrosynthetic process is iteratively employed until all $W$ belong to the building block dataset.

Currently, researchers are focusing on multi-step retrosynthetic route planning [4]. This involves a combination of single-step retrosynthetic models based on transformer or convolutional neural network, along with a multi-step Monte Carlo tree search (MCTS) framework. This approach has emerged as a valuable asset in the field of chemistry. It empowers chemists to expedite the synthesis of novel compounds with enhanced efficiency and precision, opening new vistas of exploration and discovery in the vast realm of chemical synthesis.

Among multi-step search frameworks, MCTS is a decision-making algorithm initially applied to multi-step retrosynthesis by Marwin et al. [1]. They used a neural network for single-step retrosynthesis, which is an end-to-end sequence generation effort, and MCTS for route search effort. The neural network was trained on a large dataset of known organic reactions to predict the outcome of a given chemical reaction. MCTS was used to search for the optimal synthetic route by exploring the possible chemical transformations that could be applied to the target molecule, evaluating the likelihood of each transformation using the neural network and selecting the best one to apply. The advantage of this approach is that it can efficiently explore a large search space and find a good solution, while the neural network can provide accurate predictions of chemical reactions. Lin et al. [3] also applied MCTS with a heuristic scoring function for multi-step retrosynthetic route planning. MCTS is a powerful algorithm for decision making in complex problems, however, its effectiveness and efficiency depend on several factors, such as the complexity of the problem, the branching factor of the tree, the number of simulations required, and the computational resources available. Therefore, researchers must carefully design the algorithm and its parameters to achieve the desired results. While MCTS has many advantages in decision-making problems with a large number of possible actions and states, there are also a few drawbacks to consider. Firstly, it may not work well for games or problems that have specific structures or constraints. Secondly, the search efficiency of MCTS is not very high due to its requirement of a large search space during the exploration process. Lastly, MCTS can be computationally expensive, particularly when the branching factor of the tree is high or when a large number of simulations are needed. MCTS can be parallelized [5], however, it is not commonly used in the context of retrosynthetic route planning.

Although MCTS has performed well on multi-step retrosynthesis from a decision-making point of view, its search efficiency is compromised by its large search space. Therefore, this paper proposes a method for retrosynthetic route planning using evolutionary optimization, which can purposefully find solutions while adhering to the constraints of the objective function. Secondly, the method defines the search space and limits the search scope, thereby reducing the generation of infeasible solutions. Thirdly, the proposed method employs parallel computation to reduce the search time and improve search efficiency and the number of search routes. The contributions of this paper can be summarized as follows:

❏ Evolutionary Algorithm (EA) is used for solving retrosynthetic problem, marking the first use of EA in the field of multi-step retrosynthesis.

❏ The single-step model is utilized to expand the tree nodes in the process of retrosynthetic route planning. The proposed method greatly reduces the single-step model calls, decreasing the frequency of generating invalid solutions, and improving search efficiency.

❏ Since each individual in the population is independent of each other, a parallel-EA is implemented to improve the search efficiency during the searching process.

❏ Our proposed method is executed on four case products, and performs better than Monte Carlo tree search method.

The structure of the subsequent chapters is as follows. Section II provides an overview of single-step retrosynthesis, multi-step retrosynthesis, and evolutionary algorithms. In Section III, the proposed method is presented in detail, followed by a series of comparative experiments in Section IV. The final Section concludes the paper by summarizing its key findings and contributions.

## II. Related Work

This section describes three aspects, namely, single-step retrosynthesis, multi-step search process, and evolutionary optimization. The single-step retrosynthesis is analyzed from three categories. The multi-step search process describes some state-of-the-art methods. Evolutionary optimization introduces some common genetic operators, classical probabilistic models, and some practical applications.

### A. Single-Step Retrosynthesis
Single-step retrosynthesis can be categorized into template-based, template-free, and semi-template methods. Template-based retrosynthesis [6] employs a pre-defined set of reaction templates to guide retrosynthetic analysis. These templates are based on frequently used organic synthesis reactions and can suggest possible starting materials for a desired target molecule. For instance, if the target molecule has a carbonyl group, the template-based approach may recommend using a nucleophilic addition reaction to introduce the carbonyl group and then working backwards to identify the precursor molecules for the reaction. Although this approach yields high accuracy, it requires a considerable amount of computation. Furthermore, the rule cannot cover the response range completely, and its scalability is limited. Template-based retrosynthetic reactions utilize reaction rules that may lead to reactivity conflicts. To deal with this, deep neural networks [7] have been used to resolve this problem. Computer-assisted synthesis planning (CASP) [8] was also gaining attention and molecular similarity has been found to be an effective metric for proposing and ranking single-step retrosynthetic disconnections. Deep learning-based method [9], which combine local reactivity and global attention mechanisms, was commonly used for retrosynthetic reaction prediction.

Template-free retrosynthesis [10] is a more flexible approach that does not rely on pre-defined reaction templates. Instead, it involves breaking down the target molecule into smaller fragments based on the functional groups present and then considering possible ways to connect these fragments. This open-ended approach can be useful for exploring a wider range of possible synthetic pathways. However, the accuracy of this approach is not high, especially when the reaction type is unknown, although its scalability is good. Template-free methods, on the other hand, have high scalability but lower accuracy compared to template-based methods. Retroformer [11] used a novel differentiable MCTS method to directly predict products from reactants, achieving end-to-end retrosynthesis. RetroPrime [12] combined global and local contextual information, employed a multi-layer transformer architecture, and used a multi-task learning strategy to improve accuracy and robustness of retrosynthesis prediction. Beside transformer, graph-based truncated attention (GTA) model [13] employed a truncated attention mechanism to handle interactions between different nodes and combined a reaction library with graph neural networks for retrosynthetic prediction.

Several semi-template-based approaches also exist, such as a molecular graph-enhanced transformer model [14] that represented molecular structures in graph form and used graph-based self-attention mechanisms to handle interactions between different atoms in the molecule. Additionally, the model used an adaptive structural embedding method to enhance its ability to model molecular structures.

### B. Multi-Step Search Process
Multi-step retrosynthetic reaction involves a series of consecutive single-step retrosynthesis to gradually break down the target molecule, starting from the target molecule and undergoing transformations of multiple intermediate compounds, ultimately yielding simpler starting materials. It emphasizes the requirement for multiple reaction steps to accomplish the decomposition of the target molecule, with each step serving as a crucial component of the retrosynthesis.

The state-of-the-art multi-step retrosynthetic methods include beam search, A★ algorithm, MCTS, and so on. Schwaller et al. [15] introduced a single-step retrosynthetic model predicting reactants, and the optimal synthetic pathway was found through a beam search. Meanwhile, Retro★ [16], based on A★ algorithm, combined neural networks to generate synthetic pathways efficiently. In addition to these two methods, there is another method based on MCTS search, a computational framework [17], integrating a reaction database, a generative model, and an evaluation function, was developed to guide the MCTS-based search for green synthetic pathways. The Reinforcement Learning (RL) algorithm was employed to learn from previous MCTS searches and update the evaluation function to improve search efficiency. Game tree search [18] generated new drug molecules by searching for potential reaction paths. This method combined traditional rule-based with deep learning methods and used a reinforcement learning algorithm to guide the drug molecule construction process. Another method is the one proposed by Klucznik et al. [19], which planned the synthesis path and predicted possible side reactions based on the structural characteristics and synthetic difficulty of the target compounds. However, this approach had poor performance in terms of speed of synthesis. To deal with this issue, a computer program [2] was developed to predict the optimal reaction conditions for target compounds, which was then used to control the robotic platform to perform multiple reactions simultaneously in flow reactors.

Among the above methods, beam search and A★ algorithm are relatively minority algorithms. While deep learning and reinforcement learning are more effective in solving retrosynthetic problems, they are not particularly suitable due to the nature of the algorithm. MCTS [1] stands out on retrosynthetic problems, especially for such single-step model and multi-step search combinations. Therefore, MCTS is a more general and representative retrosynthetic route search algorithm. These methods demonstrate the potential of computer-assisted synthesis planning in improving the efficiency and accuracy of chemical reactions.

### C. Evolutionary Optimization
Evolutionary algorithm (EA) is a kind of population-based optimization algorithm, through the selection of the fittest among individuals to select offspring, so as to deal with some more complex optimization problems. Evolutionary algorithms can be broadly divided into three categories: genetic based, probabilistic model-based, and individual coding based approaches. The first category includes genetic algorithm (GA) [20], differential evolution (DE) [21], and particle swarm optimization (PSO) [22]. The second one is a kind of probabilistic model-based approaches. Yao et al. [23] used Binary Genetic Algorithm (BGA) to the randomized algorithm to generate the initial population. During the

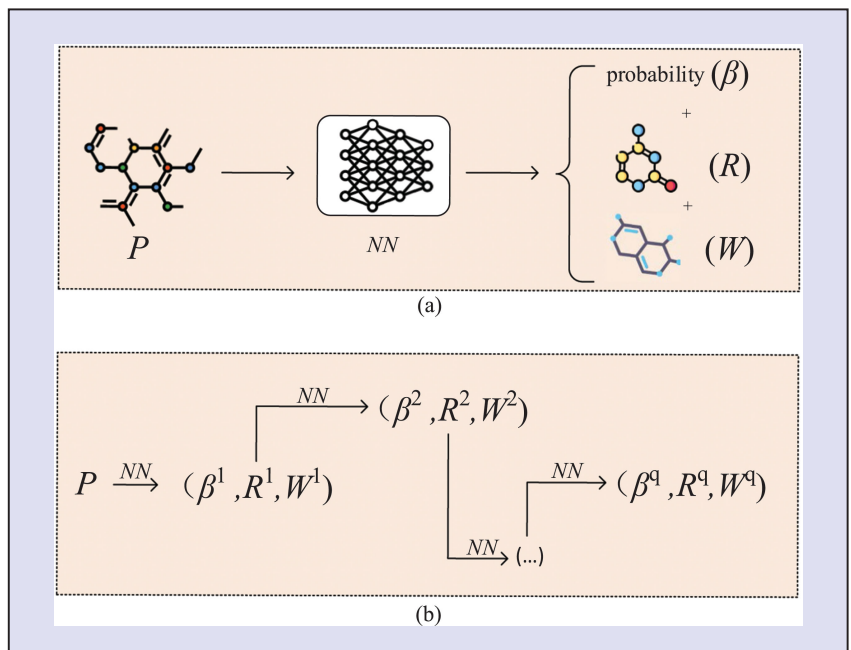

(a)

(b)

**FIGURE 2** (a) represents a single-step retrosynthetic process. In this context, $P$ denotes the target product, $NN$ expresses single-step model, $R$ and $W$ represent the reactants, where $W \in \Psi$, $\Psi$ represents building block dataset. $\beta$ signifies the probability of the reactant being involved in the process. (b) represents a multi-step retrosynthetic process, which repeats the single-step retrosynthetic process $q$ times.

coding, tree coding, graph coding and so on. Other approaches include evolution strategies (ES) [26], evolutionary programming (EP) [27], [28] and genetic programming (GP) [29], and estimation of distribution algorithms (EDA) [30].

EDA uses machine learning methods to generate new solutions and can converge to the global optimum faster than traditional EA frameworks. It can be classified into three types: univariate EDA, multi-variate EDA, and multi-objective EDA. Population-Based Incremental Learning (PBIL) [31] is a representative univariate EDA that combines genetic search–based function optimization and competitive learning. It uses a probabilistic model to generate candidate solutions and encourages diversity by introducing random perturbations. EDA/LS [32], a kind of multi-variate EDA method, samples new solutions by constructing a probabilistic model as a genetic operator. This method can assign a higher generation probability in the dominant region and accelerate the algorithm convergence. CMA-ES [33] is also a very classical multi-variate method using probabilistic model for sampling, and also a common probabilistic gradientless optimization method. The proposed method chooses EDA/LS (referred to EDA in the following optimization process of the algorithm, Bayesian network was used to do a local search and a depth-first algorithm was used to break the loop. Lee et al. [24] presented a new approach for learning the structure of Bayesian networks using a dual genetic algorithm. This method considered both the topology and ordering of BN nodes, expanding the solution space. Corriveau et al. [25] proposed a Bayesian network–based adaptive framework, BNGA, for addressing parameter setting dependencies in a steady-state genetic algorithm. The third type is based on individual coding, such as binary coding, real coding, string context) as the search framework. Pelikan et al. [34] proposed a probabilistic modeling approach, a kind of multivariate EDA method, based on Bayesian networks to solve the linkage problem and estimate the joint probability distribution of variables in a high-dimensional search space. BMDA [35] is another well-known multivariate EDA method that estimates the joint probability distribution of decision variables and uses this information to guide the search for the global optimum. Naive MIDEA [36] is a multi-objective optimization algorithm that balances convergence and diversity by using a fitness assignment scheme based on the distance between solutions. It also incorporates a crowding distance measure to promote diversity in the population. Recently, some regularity modeling approaches, i.e., RM-MEDA [37] and its variants [38], have been proposed to approximate the manifold structure of Pareto optimal solutions of multiobjective optimization problems. In summary, EDA has the ability to learn the problem structure and is suitable for solving real optimization problems with special situations.

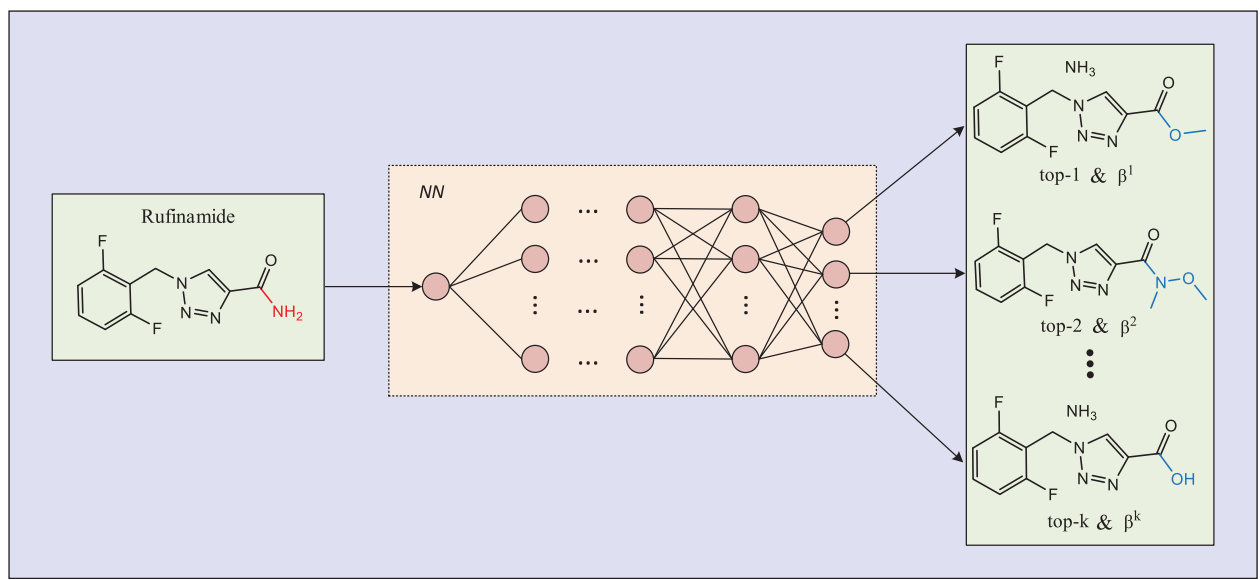

**FIGURE 3** In the framework of single-step retrosynthetic model, the top-k predicted reactants with the highest confidence, inferred by a neural network ($NN$), are utilized. The $\beta^i$ represents the probability of generating the corresponding product.

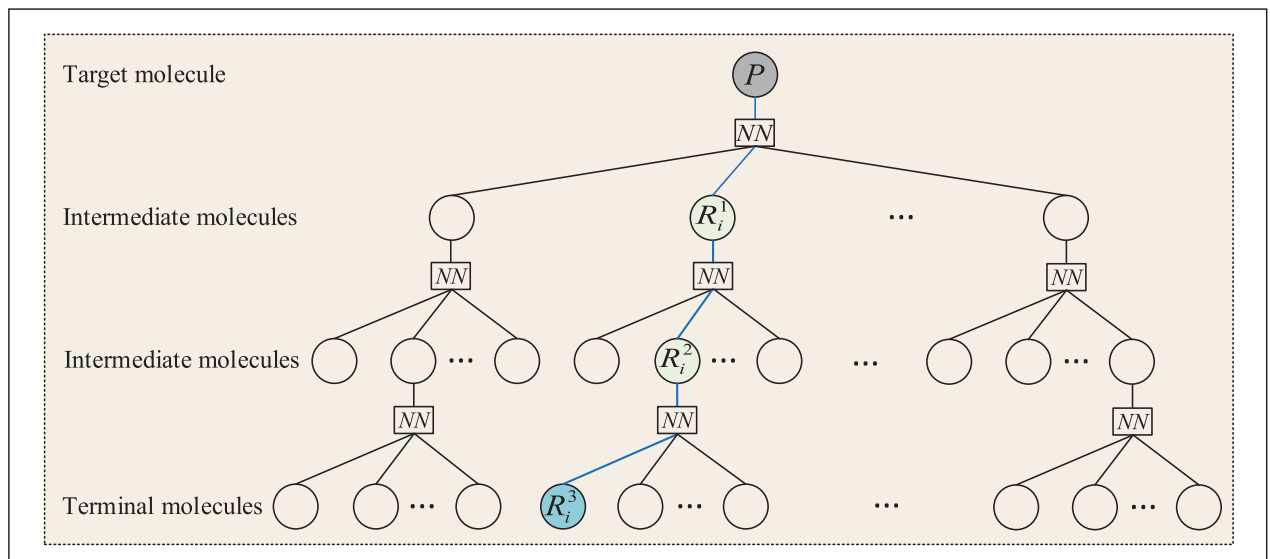

**FIGURE 4** K-cross search tree structure transformed by multi-step retrosynthesis process. Gray root node represents target product, *NN* represents transformer model, and blue leaf node stands for raw material in the building block dataset.

There are also some typical application cases [39], [40] of evolutionary optimization. For instance, In the study by Weber et al. [41], they proposed a method to process multicomponent reactions using evolutionary search methods for drug discovery. Multi-component reactions (MCRs) offer a novel method for synthesizing a wide range of compounds and compound libraries efficiently. Once considered merely a chemistry curiosity, MCRs are now acknowledged for their growing significance in drug discovery and optimization. This evolutionary search method primarily emphasizes in silico filtering, but one of its limitations is the lack of integration with deep learning techniques. Another practical application, called MemPBPF, was developed by Orozco-Rosas et al. [42] using a combination of membrane computing, a pseudo-bacterial genetic algorithm, and the artificial potential field method. The genetic algorithm component was utilized to evolve the necessary parameters within the artificial potential field method. This approach enabled the algorithm to iteratively optimize the parameters and find the best settings for generating feasible and safe paths for autonomous mobile robots. Also proposed by Orozco-Rosas et al. [43] is the mem-EAPF approach, which combines membrane computing, a genetic algorithm, and the artificial potential field method for mobile robot path planning. The approach evolves parameters within delimited compartments to minimize path length. Parallel implementations demonstrate its efficiency.

After the realization of retrosynthetic route planning algorithm, researchers can continue to study from the practical application point of view, such as Visual product [44], an industrial robotic manipulator [45] and a practical issue like machine modelling [46] can be considered.

## III. Proposed Method
This section initiates by framing the retrosynthetic problem as a tree search problem, where the objective is to find a sequence of routes leading from the root node to leaf nodes. Then the evolutionary algorithm efficiently deals with the tree search problem.

### A. Retrosynthetic Problem
In retrosynthetic analysis, reactions similar to (1a) are known as forward synthesis reactions. In this equation, a complex molecule $R$ and a simple molecule $W$ act as reactants, resulting in the formation of molecule $P$. On the other hand, (1b) is termed as single-step retrosynthetic reactions, representing the reverse of the forward synthesis reaction and forming the basis of retrosynthetic analysis. The objective of retrosynthetic analysis is to identify potential precursors or starting materials for synthesizing the target molecule $P$ using these simpler molecules $W$, which are often found in nature or obtained through standard reactions. As a result, the primary focus in single-step retrosynthetic analysis is on selecting a structurally more complex molecule $R$ for further retrosynthetic analysis. The single-step retrosynthesis process can be represented as (1c),

$$R + W \rightarrow P \quad (1a)$$
$$P \rightarrow R + W \quad (1b)$$
$$P \rightarrow R \quad (1c)$$

where $R$ and $W$ represent reactants, $P$ represents product in (1a).

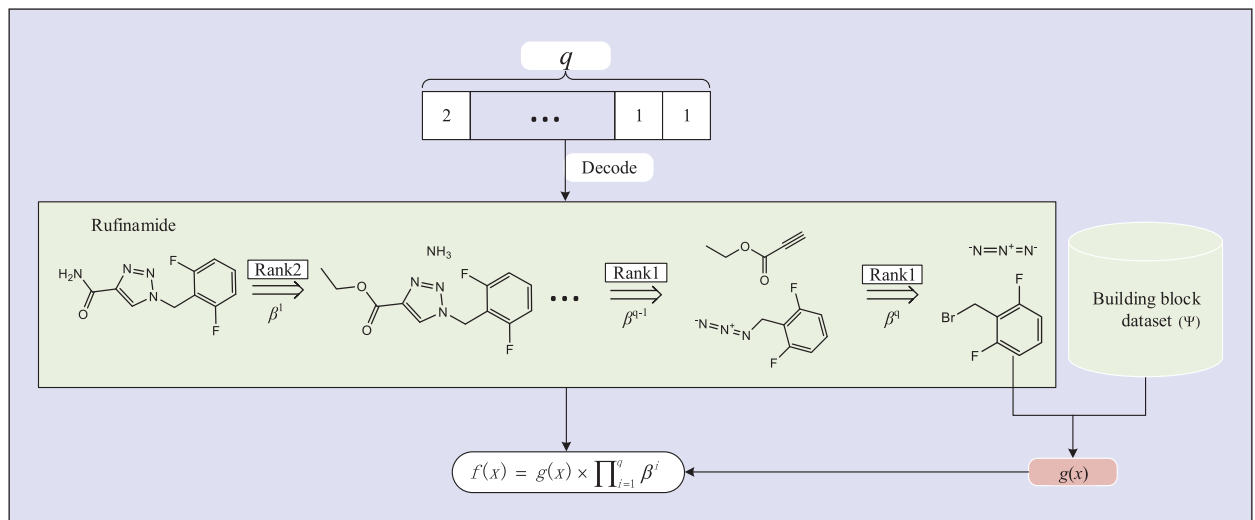

**FIGURE 5** The basic framework of objective function $f(x)$. The value of $q$ denotes the number of layers from the root node $P$ to a leaf node, as shown in Figure 4. The single-step retrosynthesis process can be interpreted as the decoding process.

For single-step retrosynthesis, the dataset is structurally expressed as $\{<P, R>\}$, where each pair contains the target product $P$ and the corresponding target reactant $R$, represented using SMILES expressions. In this problem, both the input and output are strings (SMILES expressions) [47], which can be treated as sentences. Therefore, it can be transformed into a sentence-to-sentence translation task falling under the field of Natural Language Processing (NLP). The $NN$ structural model [3] is commonly employed to address NLP challenges, with the implementation of a transformer structure in this approach. Simultaneously, Figure 2(a) shows that schematic mathematical formula is employed to describe the graph presented in Figure 1. This model takes the target product $P$ as input and generates the corresponding complex molecule $R$ and simpler molecule $W$ as output. $\beta$ signifies the probability of the reactant being involved in the process. The process is simplified in this representation, depicting a single-step retrosynthetic prediction. However, in practice, the execution of the single-step retrosynthetic model can be visualized as shown in Figure 3. In this structure, the molecule Rufinamide represents the target product $P$. The nodes top-1 to top-k represent the k most probable cases of the target product $R$ inferred by $NN$. $\beta^i$ indicates the corresponding probability of $i$-th possible case.

In multi-step retrosynthesis, the process involves a series of multiple single-step reactions, as illustrated in the simplified diagram shown in Figure 2(b). The objective of multi-step retrosynthesis is to identify one or more feasible routes from the root node to the leaf nodes. The single-step retrosynthetic model is utilized during the expansion of the tree to explore potential reactions and pathways. For example, in the context of Figure 4, the path from the root node to the leaf node, connected by the blue lines in the figure, represents a feasible route. The k-cross tree search facilitates the exploration of potential reaction pathways.

### B. Retrosynthetic Optimization Model

The number of potential chemical structures for organic molecules can reach an enormous scale, up to about $10^{60}$ compounds [48]. However, only a much smaller subset of compounds exhibits reasonable structural characteristics. Therefore, constructing a search space that explores the natural distribution of organic structures with structural rationality is a crucial challenge. The search space is defined as the top-k reactants with the highest inferred confidence in the single-step retrosynthetic reaction. The choice of the value k is crucial as it impacts the search efficiency and the quality of the solutions. If the search space is too small, there is a risk of missing the global optimal solution. Conversely, if the search space is too large, the algorithm will consume more GPU memory and time, potentially leading to local optimality and reduced efficiency. Due to the vast search space, tree search alone may not be sufficient to effectively deal with the problem. Therefore, heuristic algorithms are employed to address this limitation and provide better solutions.

To tackle the retrosynthetic problem, it is converted into an optimization problem resembling tree search. Suppose that a multi-step retrosynthesis involves $q$ single-step retrosynthesis processes. $(x, r, \beta)$ represents a multi-step retrosynthesis, where $x = (x^1, x^2, \ldots, x^q)$, and $x^i \in \{1, 2, \ldots, k\}$ represents the $x^i$-th possible output of the $i$-th single-step retrosynthesis. Let $r = (r^1, r^2, \ldots, r^q)$ denotes decoding results from $x$ using single-step retrosynthesis, as shown in Figure 3, and $r^i$ represents SMILES expression (e.g., Fc1cccc(F)c1CBr.[N-]=[N+]=[N-]) of the $i$-th single-step retrosynthesis. $\beta = (\beta^1, \beta^2, \ldots, \beta^q)$ denotes the corresponding probability of $r$. Take $(< 2, 2, 1 >, < r^1, r^2, r^3 >, < 0.8, 0.9, 0.7 >)$ as an example, there are 3 steps: in the 1st step, $r^1$ is chosen as the second reactant of NN with a probability of $\beta^1$=0.8; in the 2nd step, $r^2$ is chosen as the second reactant of NN with a probability of $\beta^2$=0.9; in the 3rd step, $r^3$ is chosen as the first reactant of NN with a probability of $\beta^3$=0.7 and it is a known molecule in $\Psi$.

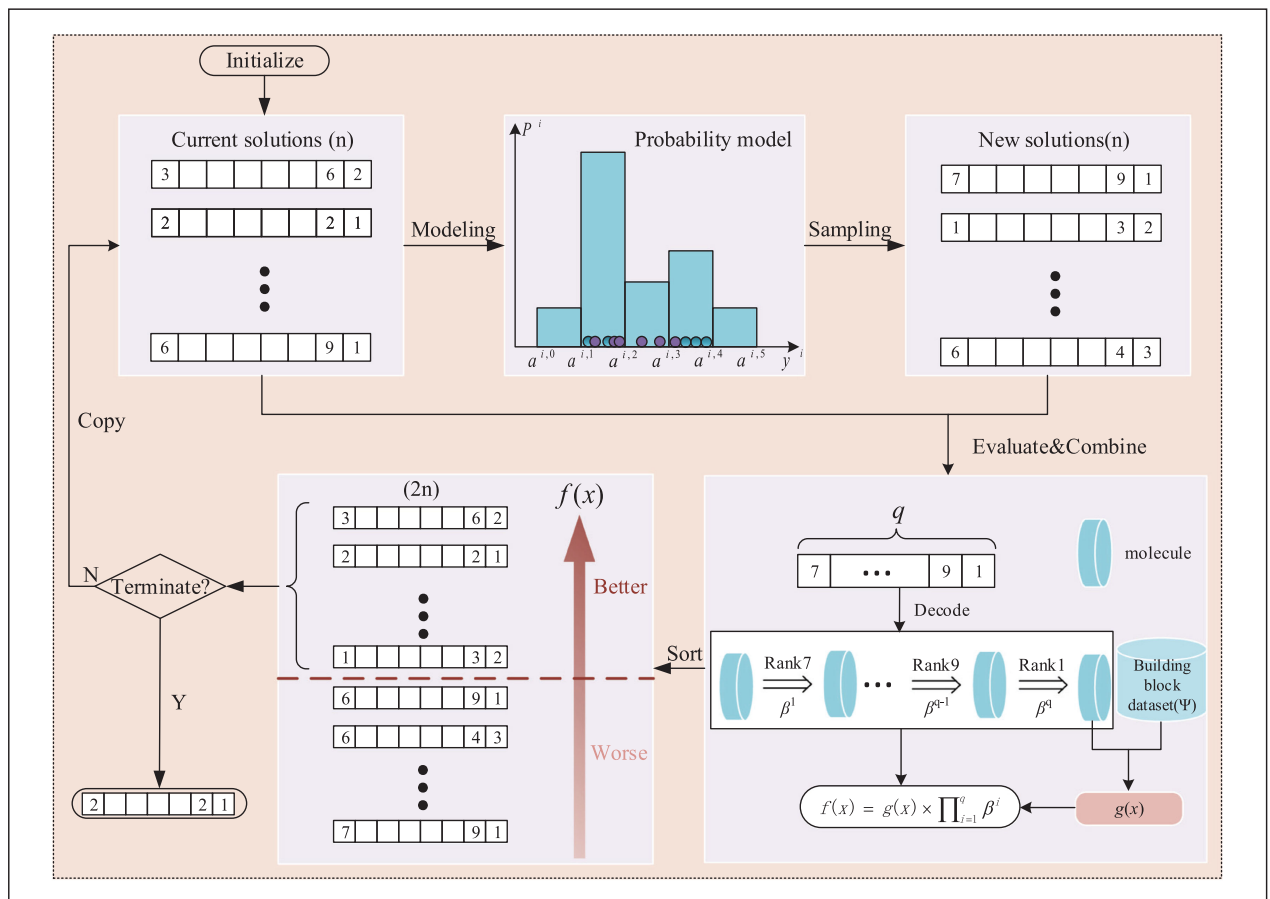

**FIGURE 6** The framework of the retrosynthetic problem utilizes Evolutionary Algorithm (EA). Within this framework, EDA operator consists of two processes: the establishment of probability model and the sampling, and it is used to generate the new solutions through sampling. The sampled results are then combined with current solutions, sorted by $f(x)$, and the first n individuals are selected as the new solutions for the next iteration.

To assess the validity of each route within the tree search problem, it is also essential to compare the molecular structure of $r^q$ with that in $\Psi$. The Morgan fingerprint [49], a widely used molecular representation, is utilized to encode the molecular structure. The Morgan fingerprint for $r^q$ and each molecule in $\Psi$ can be derived, and their molecular similarity can be calculated using RDKit [50], a popular cheminformatics library. The similarity score is defined as follows:

$$g(x) = \max_{z \in \Psi}\{\text{RDKit}(x, z)\} \qquad (2)$$

The value of $g(x)$ typically ranges between 0 and 1. In this scenario, the optimization objective shifts towards discovering the maximum value of the objective function. The mathematical expression for objective function is as follows:

$$f(x) = g(x) \times \prod_{i=1}^{q} \beta^i \qquad (3)$$

---

**Algorithm 1. Sample from Probability Model**

1: **Input**: probability model $P(y)$.
2: **Output**: new candidate solution $y$.
3: **for** $i = 1, \ldots, n$ **do**
4:    Select a bin $m$ according to probability model $P^{i,j}, j = 1, \ldots, M$.
5:    Randomly select a value $y^i$ from $[a^{i, M-1}, a^{i, M}]$ if $m = M$ or $[a^{i, m-1}, a^{i, m}]$ if $m < M$.
6: **end**
7: **Return** $y = (y^1, y^2, \ldots, y^q)$.

---

The process of obtaining the objective function is depicted in Figure 5.

### C. Evolutionary Optimization

In this paper, the proposed method adopts EA as the approach for algorithmic optimization. Prior to utilizing this algorithm, a search space transformation is required, wherein the problem search space is mapped to the variation space. Subsequently, variation, sampling and selection are performed.

#### 1) Search Space Transformation

The EA algorithm employs continuous real numbers for its coding, which necessitates mapping the continuous variation space, denoted as $\gamma$, to the discrete search space, denoted as $x$. Then, (3) is used to compute $f(x)$ for evaluating solutions. The variation space of EA is defined as $\gamma = (\gamma^1, \gamma^2, \ldots, \gamma^d)$, where $\gamma^i \in [0,1)$. When the variation space is equally divided into k segments ranging from 0 to 1, $x^i$ can be derived from $\gamma^i$ using the following mapping:

$$x^i = \lambda \quad \text{if } \frac{\lambda-1}{k} \leq \gamma^i$$
$$< \frac{\lambda}{k} \quad \lambda \in \{1, 2, \ldots, k\} \tag{4}$$

#### 2) Variation

The retrosynthetic problem is tackled using the Estimation of Distribution Algorithm (EDA) [32] during the execution of the EA process. EDA operator stands out from other operators as it generates new solutions by sampling from a histogram probabilistic model. This characteristic enables EDA to produce better or more similar solutions and converge faster. The basic framework of the proposed method is depicted in Figure 6. The process of variation comprises the establishment of the probability model and sample. To construct the probability model, the variation space $[d^{i,0}, d^{i,M}]$ for the $i$-th variable is divided into M bins, where $d^{i,0} = 0$ and $d^{i,M} = 1$ are the boundaries of the bins. Subsequently, $d^{i,1}$ and $d^{i,M-1}$

can be set as follows:

$$d^{i,1} = \max\{\gamma_1^{i,\min} - 0.5(\gamma_2^{i,\min} - \gamma_1^{i,\min}), d^{i,0}\}$$
$$d^{i,M-1} = \min\{\gamma_1^{i,\max} + 0.5(\gamma_1^{i,\max} - \gamma_2^{i,\max}), d^{i,M}\} \tag{5}$$

where $\gamma_1^{i,\min}$ and $\gamma_2^{i,\min}$ are the first and second minimum values, respectively, of the $i$-th element among the individuals in the population. Similarly, $\gamma_1^{i,\max}$ and $\gamma_2^{i,\max}$ represent the first and second maximum values, respectively, of the $i$-th element among the individuals in the population. The M-2 middle bins are of equal width, with the same size:

$$d^{i,m} - d^{i,m-1} = \frac{1}{M-2}\left(d^{i,M-1} - d^{i,1}\right) \tag{6}$$

The values assigned to each bin depend on the number of solutions found within their respective intervals, with lower values assigned to the first and last bins. To ensure that each bin has a chance of being searched, $C^{i,m}$ is used to represent the number of individuals in the $m$-th bin for variable $\gamma^i$.

$$C^{i,m} = \begin{cases} C^{i,m} + 1 & \text{if } 1 < m < M \\ 0.1 & \text{if } m = 1, M, \text{ and } d^{i,m} > d^{i,m-1} \\ 0 & \text{if } m = 1, M, \text{ and } d^{i,m} = d^{i,m-1} \end{cases} \tag{7}$$

Then the probability model can be constructed as follows:

$$P^{i,m} = \frac{C^{i,m}}{\sum_{j=1}^{M} C^{i,j}}. \tag{8}$$

The process of sample is presented in Algorithm 1.

#### 3) Decoding and Evaluate

Once new solutions are generated, they are combined with the current solutions. The $\gamma$-space is transformed into the $x$-space through search space transformation, followed by decoding to obtain $r$, which allows for the calculation of $f(x)$. The top-n individuals are then chosen based on $f(x)$ as the solutions for the subsequent iteration. These steps are repeated until the predetermined stopping criteria are met, such as reaching the maximum number of iterations or fulfilling the convergence condition.

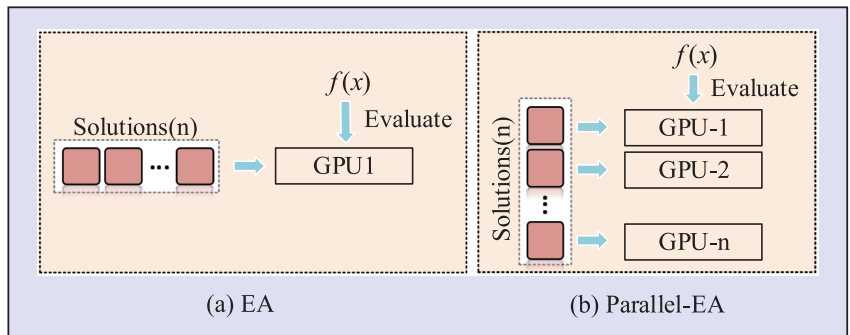

**FIGURE 7** Parallel computing of EA. The population consists of $n$ individuals, represented by red squares. In the standard EA, each individual is evaluated sequentially. However, in parallel EA, different individuals can be assigned to separate GPUs for evaluation simultaneously.

**TABLE I** Descriptions of 10 reaction classes and the ratio of USPTO_50K and USPTO_MIT.

| REACTION CLASS | REACTION NAME | USPTO_50K(%) | USPTO_MIT(%) |
|---|---|---|---|
| 1 | Heteroatom alkylation and arylation | 30.3 | 29.9 |
| 2 | Acylation and related processes | 23.8 | 24.9 |
| 3 | C-C bond formation | 11.3 | 13.4 |
| 4 | Heterocycle formation | 1.8 | 0.7 |
| 5 | Protections | 1.3 | 0.3 |
| 6 | Deprotections | 16.5 | 14.1 |
| 7 | Reductions | 9.2 | 9.4 |
| 8 | Oxidations | 1.6 | 2.0 |
| 9 | Functional group interconversion (FGI) | 3.7 | 5.0 |
| 10 | Functional group addition (FGA) | 0.5 | 0.2 |

**TABLE II** Analysis of the number (mean(std.dev.))[percentage] of calling single-step model for MCTS and EA. Percentage in "[.]" expresses increase or decrease of EA compared with MCTS in the same column and under the same product over 30 independent runs.

| PRODUCTS | METHOD | ONE SOLUTION | TWO SOLUTIONS | THREE SOLUTIONS |
|---|---|---|---|---|
| DemoA | MCTS | 1.20e+2(8.34e+1) | 1.95e+2(8.34e+1) | 2.74e+2(1.12e+2) |
| DemoB | MCTS | 3.81e+2(3.13e+2) | 4.56e+2(3.13e+2) | 4.89e+2(3.39e+2) |
| DemoC | MCTS | 2.13e+3(1.70e+3) | 3.38e+3(2.19e+3) | 3.65e+3(2.35e+3) |
| DemoD | MCTS | 4.58e+3(2.77e+3) | 4.87e+3(2.07e+3) | 4.82e+3(2.00e+3) |
| DemoA | EA | 2.30e+1(1.89e+1)[80.8%↓] | 5.85e+1(4.19e+1)[70.0%↓] | 8.23e+1(4.45e+1)[70.0%↓] |
| DemoB | EA | 2.48e+1(2.33e+1)[93.5%↓] | 6.64e+1(3.87e+1)[85.4%↓] | 1.11e+2(5.48e+1)[77.3%↓] |
| DemoC | EA | 2.60e+1(1.45e+1)[98.8%↓] | 2.11e+3(1.28e+3)[35.8%↓] | 2.41e+3(3.28e+2)[34.0%↓] |
| DemoD | EA | 2.69e+3(1.11e+3)[41.3%↓] | 2.69e+3(1.01e+3)[44.8%↓] | 3.16e+3(3.45e+2)[34.4%↓] |

**4) Parallel Implementation**
EA is a population-based search method, where each individual in the population is independent and represents a complete retrosynthetic route, making it natural for parallel computation. Leveraging this advantage, parallel EA distributes $n$ individuals across different GPUs for simultaneous computation, following a specific order:

$$s = j\%m \qquad (9)$$

where $j$ represents the $j$-th individual in the population, $m$ is the number of GPU, then $s$ represents the remainder of dividing $j$ by $m$. In this case, the $j$-th individual is assigned to the $s$-th GPU for computation, as shown in Figure 7.

**TABLE III** Analysis of search capability (mean(std.dev.))[percentage] on beam size top-10, top-15, top-20. Percentage in "[.]" expresses increase or decrease of EA compared with MCTS in the same column and under the same product. Data in the table is over 30 independent runs.

(a) Num of calling single-step model based on searching three feasible solutions.

| PRODUCTS | METHOD | TOP-10 (NUM OF CALLS) | TOP-15 (NUM OF CALLS) | TOP-20 (NUM OF CALLS) |
|---|---|---|---|---|
| DemoA | MCTS | 2.74e+2(1.12e+2) | 7.76e+2(3.72e+2) | 9.21e+2(6.06e+2) |
| DemoB | MCTS | 4.89e+2(3.39e+2) | 7.02e+2(3.84e+2) | 8.88e+2(2.29e+2) |
| DemoC | MCTS | 3.65e+3(2.35e+3) | 3.99e+3(1.98e+3) | 8.00e+3(4.56e+3) |
| DemoD | MCTS | 4.82e+3(2.00e+3) | 7.29e+3(2.55e+3) | 2.65e+4(3.53e+3) |
| DemoA | EA | 8.23e+1(4.45e+1)[70.0%↓] | 4.50e+2(1.18e+2)[42.0%↓] | 9.62e+2(2.74e+2)[4.4%↓] |
| DemoB | EA | 1.11e+2(5.48e+1)[77.3%↓] | 3.96e+2(1.58e+2)[43.6%↓] | 9.05e+2(3.95e+2)[1.9%↓] |
| DemoC | EA | 2.41e+3(3.28e+2)[34.0%↓] | 3.45e+3(6.26e+2)[13.5%↓] | 7.92e+3(2.16e+3)[1.0%↓] |
| DemoD | EA | 3.16e+3(3.45e+2)[34.4%↓] | 6.47e+3(1.70e+3)[11.2%↓] | 1.06e+4(1.46e+3)[60.0%↓] |

(b) Search time (sec) based on searching three feasible solutions.

| METHODS | DemoA (TOP-10) | DemoB (TOP-10) | DemoC (TOP-10) | DemoD (TOP-10) |
|---|---|---|---|---|
| MCTS | 1.63e+3(9.48e+2) | 3.45e+3(3.23e+3) | 1.79e+4(1.18e+4) | 2.85e+4(1.12e+4) |
| EA | 1.76e+2(5.92e+1)[89.2%↓] | 3.30e+2(1.54e+2)[90.4%↓] | 4.83e+3(4.50e+2)[73.0%↓] | 4.85e+3(1.79e+3)[83.0%↓] |

(c) Num of feasible search routes based on 200 iterations.

| PRODUCTS | METHOD | TOP-10 (NUM OF RESULTS) | TOP-15 (NUM OF RESULTS) | TOP-20 (NUM OF RESULTS) |
|---|---|---|---|---|
| DemoA | MCTS | 4.00e+0(2.23e+0) | 3.27e+0(1.09e+0) | 2.66e+0(1.88e+0) |
| DemoB | MCTS | 3.90e+0(0.55e+0) | 3.86e+0(0.69e+0) | 3.20e+0(0.92e+0) |
| DemoC | MCTS | 3.46e+0(1.49e+0) | 3.44e+0(1.16e+0) | 2.88e+0(0.73e+0) |
| DemoD | MCTS | 3.46e+0(2.36e+0) | 2.60e+0(1.96e+0) | 1.66e+0(0.94e+0) |
| DemoA | EA | 9.53e+0(3.15e+0)[138%↑] | 3.86e+0(1.01e+0)[18.0%↑] | 2.71e+0(1.67e+0)[1.9%↑] |
| DemoB | EA | 4.33e+0(1.69e+0)[11%↑] | 4.13e+0(0.54e+0)[7.0%↑] | 3.87e+0(0.73e+0)[20.9%↑] |
| DemoC | EA | 4.23e+0(1.24e+0)[22%↑] | 3.68e+0(1.03e+0)[7.0%↑] | 3.14e+0(0.75e+0)[9.0%↑] |
| DemoD | EA | 3.60e+0(2.57e+0)[4.6%↑] | 3.12e+0(2.61e+0)[20.0%↑] | 1.73e+0(1.06e+0)[4.2%↑] |

## IV. Experiment

This section starts by introducing the datasets and parameter settings. It then conducts a comparative study using various methods and performs extensive experiments on four case products. Finally, several charts are provided along with corresponding analyses.

### A. Datasets

This experiment involved retrosynthetic route planning for four case products. Two commonly used benchmark datasets, namely, USPTO_50K [51], [52] and USPTO_MIT, were utilized to train single-step retrosynthetic model. The USPTO_50 K dataset was extracted from the patent literature of the United States, while USPTO_MIT was previously employed by Lin et al. [3]. The distribution of reaction classes in the two datasets is presented in Table I. To define the terminal nodes or reactants, the building block dataset [3] is built, which contains 93563 commercially available molecules.

### B. Parameters Setting

During the experiment, the parameters are outlined below:

❏ Population size. Set $N$ = 42. Larger populations typically have higher diversity, which can increase the robustness of the algorithm. But larger populations require more computational resources and time to evaluate and evolve each generation.

❏ Thread pool size. Set $P$ = 42, which is equal to population size.

❏ Termination condition. The population converges in about 150 generations at most; therefore, the search terminated after 200 iterations.

❏ Number of bins. Set $M$ = 10. $M$ represents the number of bins in each dimension of the population, which ranges from 0 to 1. Because the results of the single-step model are encoded as the numbers 0-9, a total of 10 results. Smaller $M$ will miss some encoding results, while larger $M$ leads to computational redundancy.

Each experiment is executed for 30 independent runs, and the mean and variance are used for result analysis. The running program is written using python scripts. During the search process, the retrosynthetic routes are explored using 3 NVIDIA GeForce RTX 4090 GPUs.

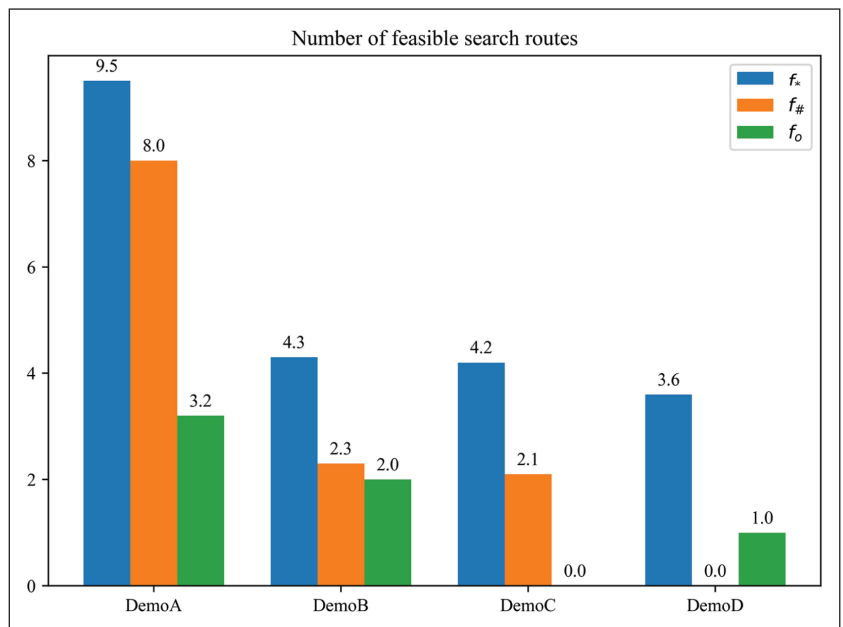

**FIGURE 8** The number of feasible search routes with different objective functions based on 200 iterations, over 30 independent runs.

### C. Comparative Experiment

The experimental section conducted a series of comparative experiments to validate the effectiveness of the proposed algorithm. In particular, it scrutinized the impact on the number of calls to the single-step model, the search capability in a larger search space, and the count of feasible search solutions. Additionally, several objective optimization functions were designed, and the algorithm's search time was tested.

#### 1) Comparison Experiment on Calling Single-Step Model

The frequency of calling single-step model directly affects both search time and search efficiency. MCTS and EA were employed to search for one, two, and three solutions for the four case products, as depicted in Table II. The values in the table indicate the average and standard deviation of the number of calling single-step model. As the number of solutions increase, both MCTS and EA require a higher count of model calls. However, EA consistently outperforms MCTS for one, two, and three solutions. Furthermore, across the four case products, the number of calling single-step model reduced by an average of 78.6%, 59.0%, and 53.9%, respectively. Consequently, EA exhibits faster search speed and higher search efficiency when compared to MCTS.

#### 2) Comparison Experiment on the Search Capability

Tests with three different beam sizes (top-10, top-15, and top-20) were conducted to compare the search capabilities of the two algorithms in different search spaces.

Firstly, with the expansion of search space, the number of calling single-step model for both MCTS and EA increases significantly. As shown in Table III(a), due to the advantage of EA search mechanism, under the beam size of top-10, top-15, and top-20, the number of calling single-step model of EA decreased by 53.9%, 27.6%, and 16.8% on average, respectively, compared to that of MCTS based on searching three feasible solutions.

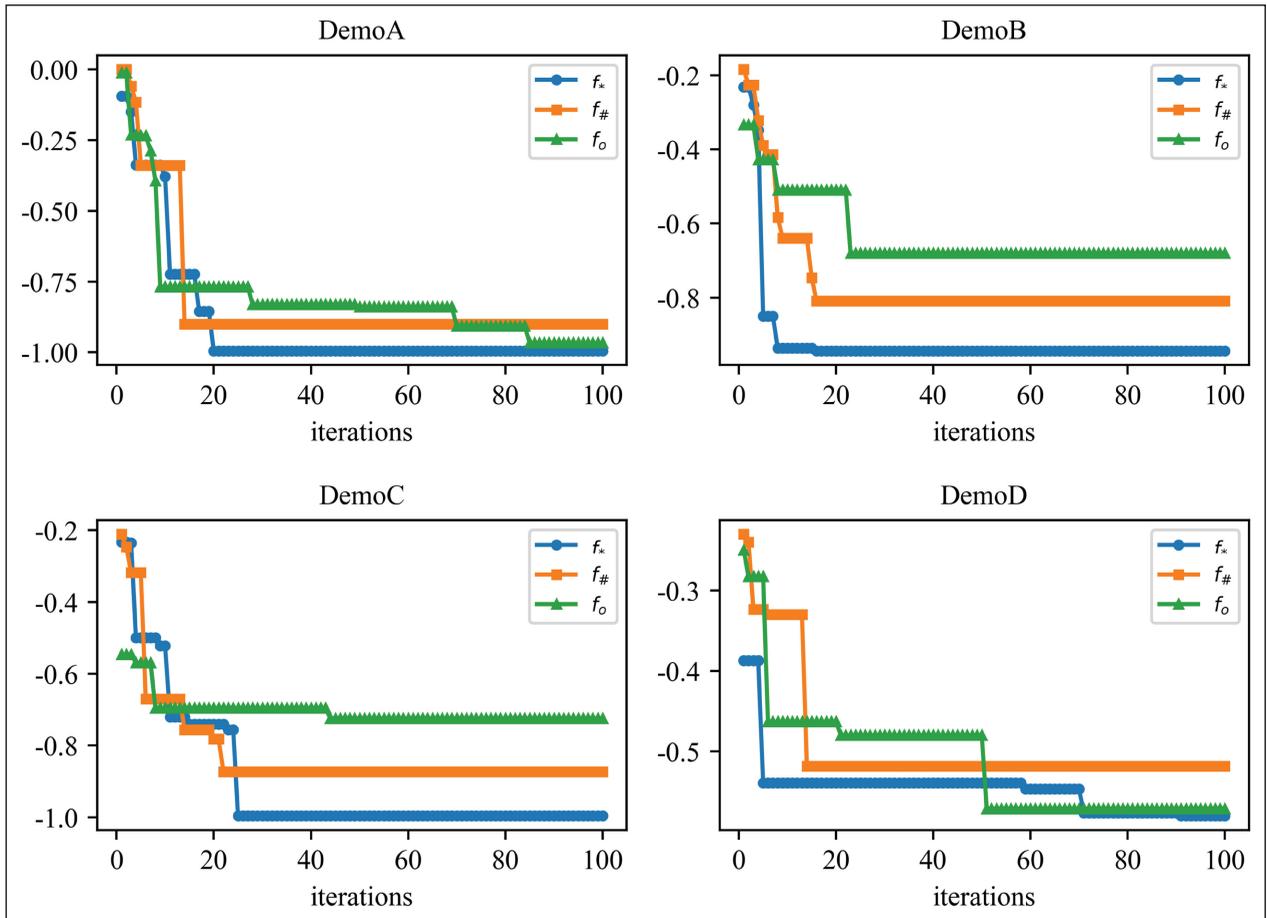

**FIGURE 9** Evolution of objective values of three different objective functions. Objective function $f_*$ outperforms the other two functions on the four target products.

Secondly, this experiment is conducted with a beam size of top-10. Notably, there is a progression in the complexity of molecular structures from DemoA to DemoD, which subsequently leads to longer search time. When focusing on the same target product, the search time for EA decreased significantly compared to that of MCTS, with an average reduction of 83.9% across the four case products based on searching three feasible solutions, as shown in Table III(b).

Thirdly, as the search space increased in Table III(c), the complexity of the search also rises, meanwhile, the number of search routes for four case products gradually decreased under the same number of iterations. However, EA outperforms MCTS in terms of the number of search routes across top-10, top-15, and top-20. Among them, the performance on top-15 and top-20 is normal, but the performance on top-10 is particularly noteworthy, with the number of search routes increasing 1.38 times on average. This could be attributed to the relatively simplistic and easily searchable molecular structure of DemoA.

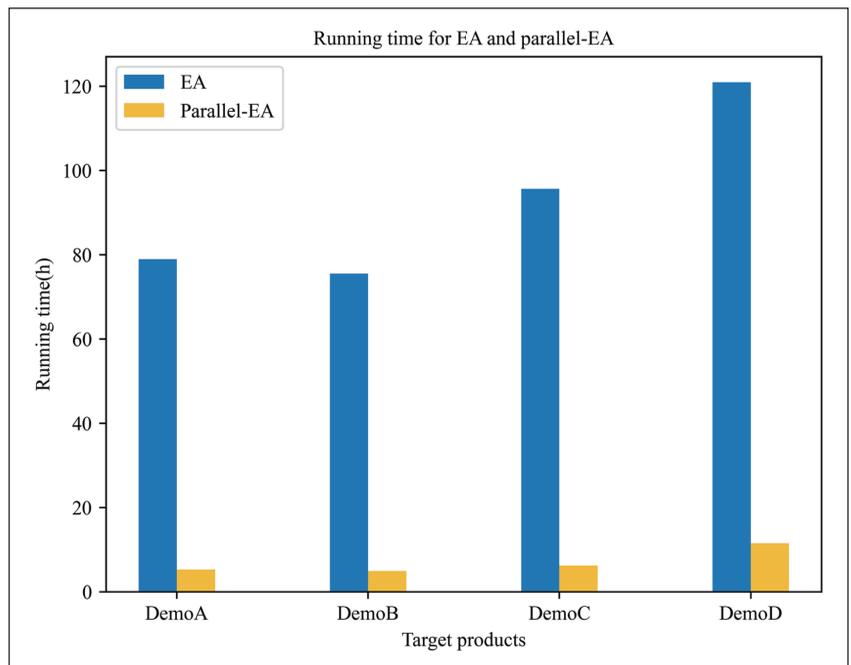

**FIGURE 10** Running time for EA and Parallel-EA. The algorithm stops at the maximum number of iterations 100.

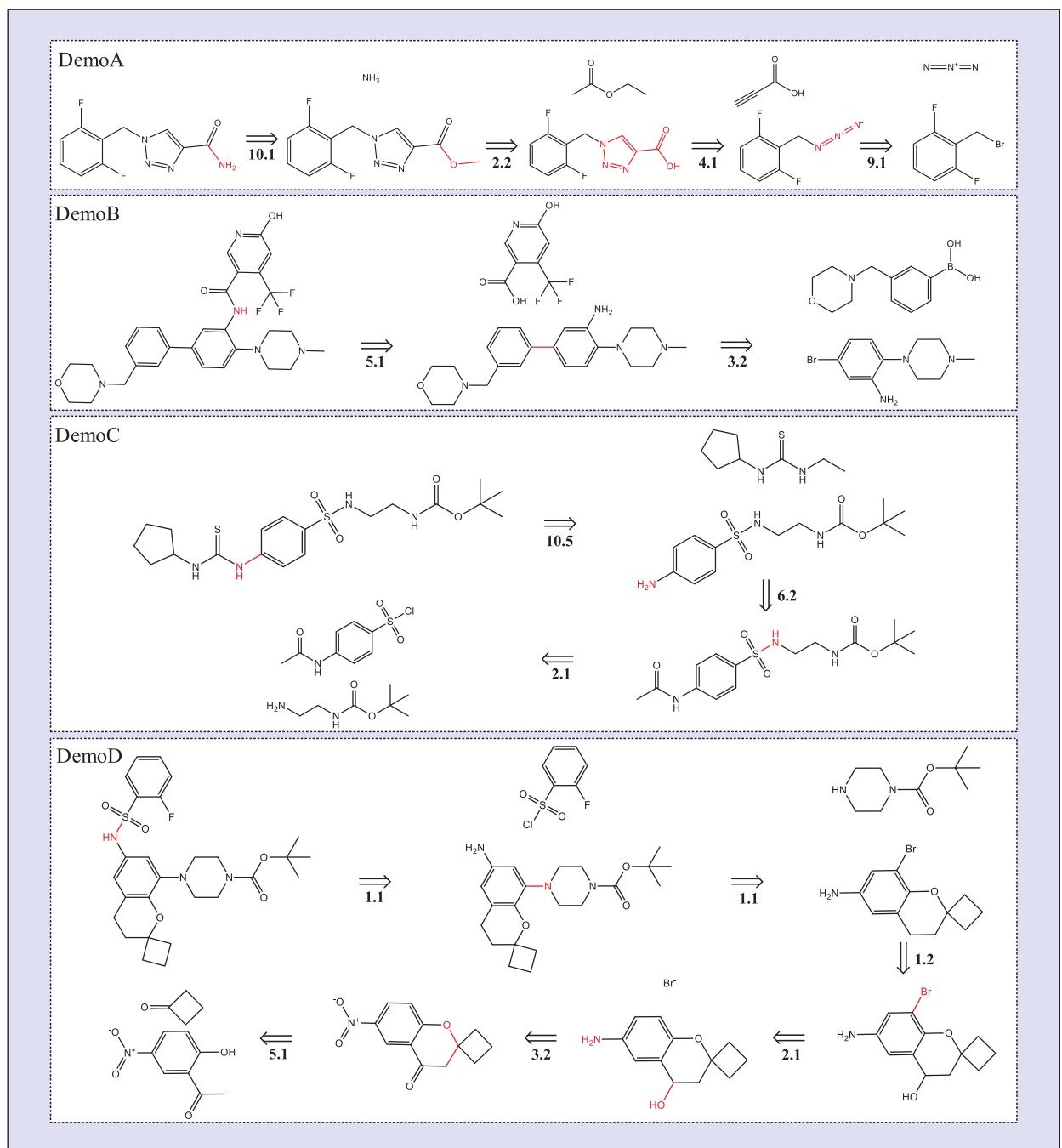

**FIGURE 11** Multi-step retrosynthetic routes. The affected functional groups in each step are marked red. The number before "." and after "." indicates the reaction class and ranking in the top-10 prediction, respectively.

In summary, across the beam size of top-10, top-15, and top-20, EA

consistently outperforms MCTS by generating a higher number of feasible search routes. Furthermore, EA demonstrates lower calling frequency of single-step model and shorter search time compared to MCTS based on searching the same feasible solutions. Overall, EA exhibits a significant improvement in search efficiency over MCTS.

**3) Comparison Experiment on Different Objective Functions**

To explore the impact of different objective functions on the convergence and the ability to search feasible solutions, three objective functions are designed. Assuming there is a population of n solutions, pop= $\{x_1, x_2, \ldots, x_n\}$, with the fitness of each solution denoted as $f(x_i) \in (0, 1)$.

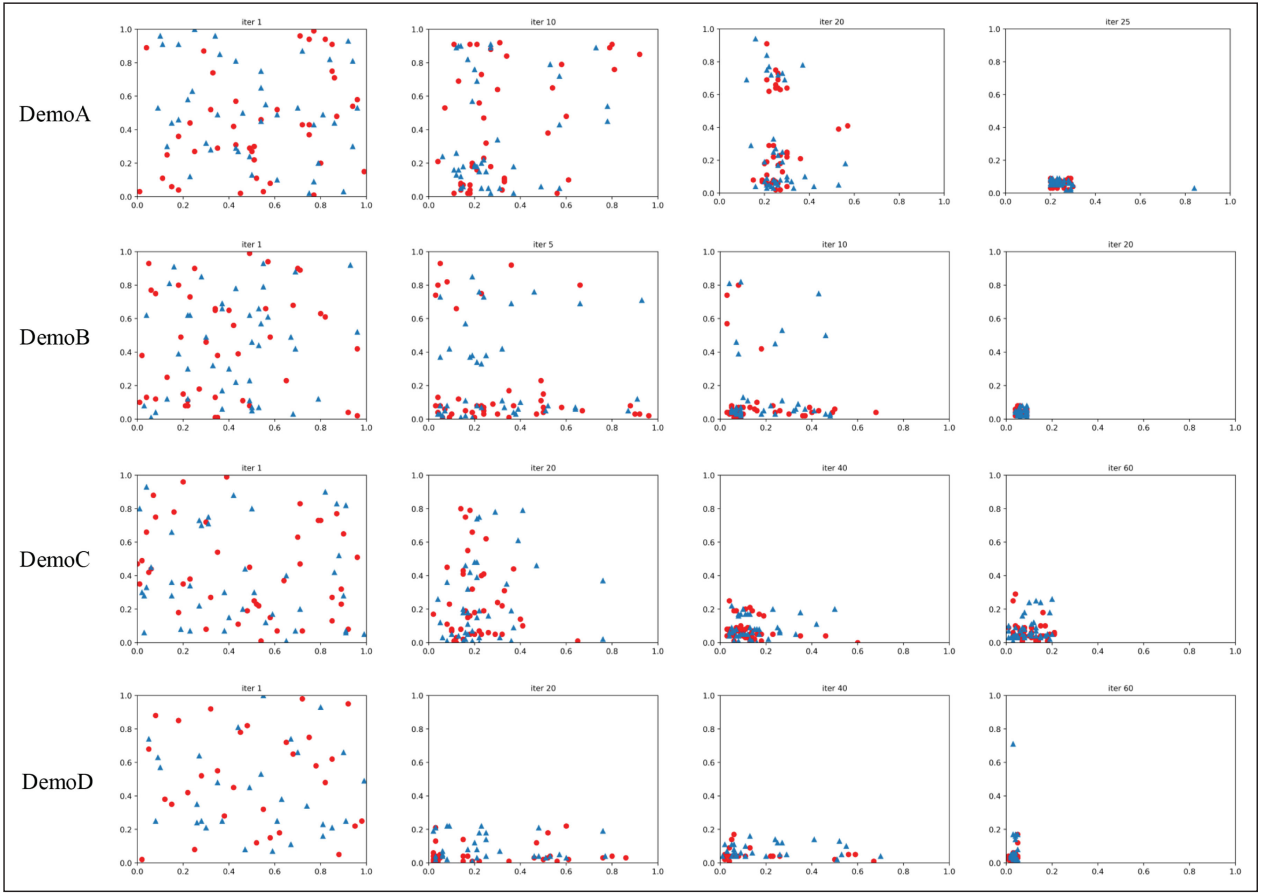

**FIGURE 12** The population distribution of four case products with the increase of the number of iterations. Red "o" stands for current population, and blue "Δ" stands for new population generated from current population.

To compare the convergence of the objective functions, they are transformed into their negative counterpart. Therefore, the fitness value or selection probability of the $i$-th individual is designed as follows:

$$f_* = -f(x_i) \tag{10}$$

or

$$f_\# = -e^{f(x_i)} \tag{11}$$

or

$$f_o = -\frac{f(x_i)}{\sum_{j=1}^{n} f(x_j)} \tag{12}$$

During the selection process, solutions are ordered based on the fitness values for $f_*$, $f_\#$, and $f_o$. $f_*$ is the negation of the original objective function, and $f_\#$ engineered to increase the separation between target values. Meanwhile, $f_o$ utilizes the roulette wheel selection sampling method, inspired by the popular game of roulette [53]. This selection mechanism assigns a higher probability of selection to individuals with higher fitness values. Consequently, individuals with superior fitness have a greater chance of being chosen, leading to their retention in the evolutionary process. By favoring the selection of better individuals, this approach enhances the convergence and optimization capabilities of the genetic algorithm. Figure 8 shows that $f_*$ yields more feasible solutions compared to the other objective functions, making it the most effective among them. To further assess the convergence of the three different objective functions on the four case products, the value range of $f_\#$ is normalized to match the same range as $f_*$ and $f_o$, which is from –1 to 0. As shown in Figure 9, the results clearly demonstrate that $f_*$ outperforms the other two objective functions on DemoA, DemoB, and DemoC. Although $f_o$ shows some similarity to $f_*$ on DemoD, $f_*$ still performs better after 80-th iteration.

### 4) Comparison Experiment on Search Time Between EA and parallel-EA

The implementation of parallel computation allowed us to expedite the running process of EA, leveraging its search mechanism effectively. The results of the final search time are depicted in Figure 10, with the algorithm stopping at 100 epochs and being tested on four case products. Remarkably, the running time of parallel-EA was found to be 10-15 times faster compared to that of ordinary EA. This substantial improvement in speed demonstrates the remarkable advantage of parallel computation in accelerating the search process of EA,

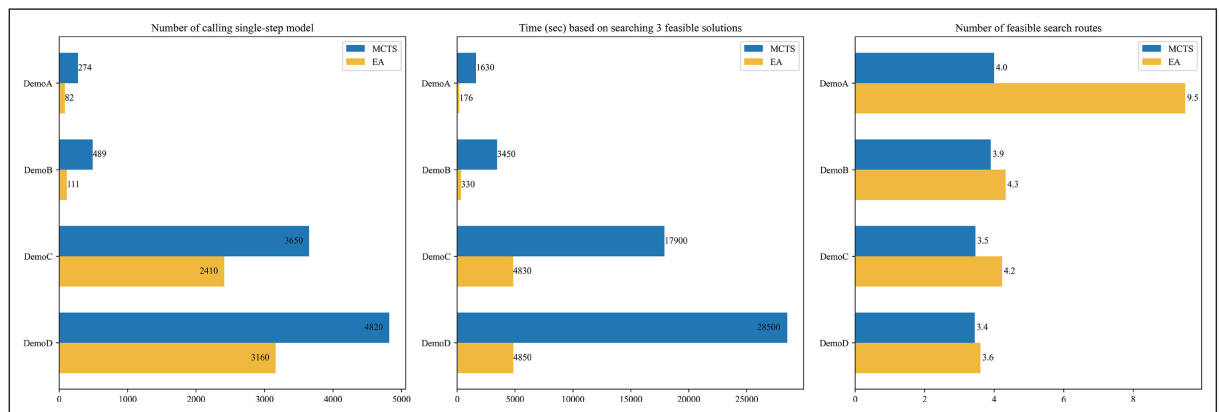

**FIGURE 13** The experiment of calling single-step model and search time is conducted based on identifying three feasible solutions. Additionally, the experiment of the number of feasible routes is carried out based on 200 iterations.

enabling more efficient exploration of the retrosynthetic search space.

### D. Visualization of Results

This section provides a comprehensive visual analysis of EA's performance in multi-step retrosynthesis. It includes the presentation of the retrosynthetic routes and the convergence of the population during the search process. Furthermore, EA's performance is compared with MCTS from three different perspectives.

In the retrosynthetic analysis, four case products were used. DemoA refers to Rufinamide [3], an antiepileptic triazole derivative. DemoB, DemoC, and DemoD are from Klucznik et al. [19], Li et al. [54], and Segler et al. [1], respectively. The retrosynthesis of these four products required 4, 2, 3, and 6 steps, respectively. As depicted in Figure 11, DemoA's search procedure comprises four steps leading to the final products being in $\Psi$. The impacted functional groups are highlighted in red. The digits before and after the "." indicate the reaction type and the substance's positioning in the results of the single-step model inference, correspondingly. These four case products have received approval from chemical experts.

During the search process, the distribution of the encoded population is visualized, as shown in Figure 12. As the iteration increases, the population distributions gradually converge to the similar position. DemoA and DemoB demonstrate nearly rapid convergence after approximately 20 iterations, which can be attributed to their relatively simple molecular structures. In contrast, DemoC and DemoD, with more intricate molecular configurations, necessitate a broader range of reaction rules for analysis, presenting greater search challenges. As a result, these compounds converge around the 60-th iteration. Despite the intricate nature of the problem and the gradual rate of convergence, all four compounds ultimately achieve convergence.

EA consistently outperforms MCTS in different aspects, as shown in Figure 13. In the context of identifying three feasible solutions, there is a considerable decrease in the number of calling single-step model and search time. Additionally, what is particularly noteworthy is EA's significant enhancement in searching feasible solutions under 200 iterations, especially for DemoA, because of its relatively simple molecular structure. Furthermore, the parallel implementation of EA results in an exponential reduction in time, making it a highly efficient approach for retrosynthetic route planning. Overall, the utilization of EA in retrosynthetic route planning demonstrates its capability to efficiently handle complex and challenging synthesis problems, thereby providing chemists with valuable and promising potential routes for target compounds.

## V. Conclusion

This work proposes a novel method for dealing with retrosynthetic route planning with Evolutionary Algorithms (EA). To be specific, a single-step model is used in the proposed method to learn the reaction rules from the datasets, and then a well-designed EA is adopted to deal with the optimization task, which is modeled from the retrosynthesis. The experiments show that the number of calling single-step model and the search time are reduced by and average of 53.9% and 83.9%, respectively. The number of search routes increases by 1.38 times. Multiple feasible routes have been successfully discovered and recognized by chemists, showcasing the practical utility of the method.

Although the method proposed in this study has demonstrated promising results, there are several directions that can be pursued:

1) Continuous encoding, used in the genetic operator, may affect the search efficiency of the algorithm. Therefore, selecting discrete encoding genetic operators can better match this research.

2) In practical applications, it may be more appropriate to model retrosynthetic route planning as a multi-objective optimization problem, especially in complex situations. By employing some promising generation paradigm [55] and solution estimation approach [56] that are specifically designed for multi-objective optimization, one can effectively address the challenges in such a problem.

3) After the implementation of the algorithm, it is necessary to continue to update some practical applications,

such as a synthesis planning tool [57]. This could help chemists synthesize molecules more efficiently.

## Acknowledgment

This work was supported in part by the National Natural Science Foundation of China under Grant 62306174, in part by China Postdoctoral Science Foundation under Grant 2023M742259 and Grant 2023TQ0213, and in part by the Postdoctoral Fellowship Program of CPSF under Grant GZC20231588.